%% file: egpaper_for_review.tex
\documentclass[10pt,twocolumn,letterpaper]{article}

\usepackage{iccv}
\usepackage{times}
\usepackage{epsfig}
\usepackage{graphicx}
\usepackage{amsmath}
\usepackage{amssymb}

\usepackage[normalem]{ulem} 
\usepackage{booktabs}
\usepackage{amsmath}
\usepackage{amssymb}
\usepackage{multirow}
\usepackage{algorithm}
\usepackage[noend]{algorithmic}
\usepackage{makecell}
\usepackage{enumitem}
\usepackage{epigraph}
\usepackage{csquotes}
\usepackage{mathtools}
\usepackage{color, colortbl}
\definecolor{LightCyan}{rgb}{0.88,1,1}
\definecolor{LightRed}{rgb}{1.0, 0.91, 0.91}
\definecolor{LightGray}{rgb}{0.88,0.88,0.88}
\definecolor{VeryLightGray}{rgb}{0.93,0.93,0.93}
\newcommand{\dashrule}[1][black]{%
  \color{#1}\rule[\dimexpr.5ex-.2pt]{4pt}{.4pt}\xleaders\hbox{\rule{4pt}{0pt}\rule[\dimexpr.5ex-.2pt]{4pt}{.4pt}}\hfill\kern0pt%
}

\usepackage[pagebackref=true,breaklinks=true,letterpaper=true,colorlinks,bookmarks=false]{hyperref}

\iccvfinalcopy 

\def\iccvPaperID{10857} 

\ificcvfinal\fi

\begin{document}

\title{Weakly Supervised Relative Spatial Reasoning for Visual Question Answering}

\author{
Pratyay Banerjee \quad Tejas Gokhale \quad Yezhou Yang \quad Chitta Baral  \\
Arizona State University \\
\texttt{\{pbanerj6, tgokhale, yz.yang, chitta\}@asu.edu}               \\
}


\maketitle
\ificcvfinal\thispagestyle{empty}\fi

\begin{abstract}
Vision-and-language (V\&L) reasoning necessitates perception of visual concepts such as objects and actions, understanding semantics and language grounding, and reasoning about the interplay between the two modalities.
One crucial aspect of visual reasoning is spatial understanding, which involves understanding relative locations of objects, i.e.\ implicitly learning the geometry of the scene.
In this work, we evaluate the faithfulness of V\&L models to such geometric understanding, by formulating the prediction of pair-wise relative locations of objects as a classification as well as a regression task.
Our findings suggest that state-of-the-art transformer-based V\&L models lack sufficient abilities to excel at this task.
Motivated by this, we design two objectives as proxies for 3D spatial reasoning (SR) -- object centroid estimation, and  relative position estimation, and train V\&L with weak supervision from off-the-shelf depth estimators.
This leads to considerable improvements in accuracy for the ``GQA'' visual question answering challenge (in fully supervised, few-shot, and O.O.D settings) as well as improvements in relative spatial reasoning.
Code and data will be released \href{https://github.com/pratyay-banerjee/weak_sup_vqa}{here}.
\end{abstract}

\input{Latex/01_introduction}
\input{Latex/02_related_work}

\input{Latex/03_relative_spatial_reasoning}

\input{Latex/04_method}

\input{Latex/tables/spatial_task_2d_3d}

\section{Experiments}
\paragraph{Datasets.}
We evaluate our methods on two popular 
benchmarks, GQA~\cite{hudson2019gqa} and  GQA-OOD~\cite{kervadec2020roses}, both of which contain spatial reasoning visual questions requiring compositionality and relations between objects present in natural non-iconic images. 
Both datasets have a common training set, but differ in the test set: GQA uses an i.i.d.\ split, while GQA-OOD contains a distribution shift.
There are 2000 unique answers in these datasets, and questions can be categorized based on the type of answer: binary (yes/no answers) and open-ended (all other answers).

\paragraph{Evaluation Metrics.}
For evaluating performance in fully-supervised, few-shot, as well as O.O.D.\ settings for the GQA task, we use metrics defined in~\cite{hudson2019gqa}.
These include exact match accuracy, accuracy on the most frequent head answer-distribution, infrequent tail answer-distribution, consistency to paraphrased questions, validity, and plausibility of spatial relations\footnote{Detailed definitions of these metrics can be found in Section 4.4.\ of Hudson~\etal~\cite{hudson2019gqa} or accessed on the  \href{https://cs.stanford.edu/people/dorarad/gqa/evaluate.html}{GQA Challenge webpage}}.
We evaluate SR tasks using mean-squared error (MSE) for SR-Regression and classification accuracy for SR bin-classification.

\paragraph{Model Architectures.} 
LXMERT contains 9 language transformer encoder layers, 5 visual layers, and 5 cross-modal layers. 
This feature extractor can be replaced by any other transformer-based V\&L model. 
Our Fusion transformer has 5 cross-modal layers with a hidden dimension of $H=512$. 
For visual feature extraction, we use ResNet-50~\cite{he2016deep} pre-trained on ImageNet~\cite{russakovsky2015imagenet} to extract image patch features, with 50\% overlap, and Faster RCNN pre-trained on Visual Genome~\cite{krishna2017visual} to extract the top 36 object features. 
We use $3\times3$, $5\times5$, $7\times7$ patches, and the entire image as the spatial image patch features. 
The image is uniformly divided into a set of overlapping patches at multiple scales.


\paragraph{Training Protocol and Hyperparameters.} 
Our Fusion transformer has 5 cross-modal layers with a hidden dimension of $H=512$. 
All models are trained for 20 epochs with a learning rate of $1\mathrm{e}{-5}$, batch size of 64, using Adam~\cite{kingma2014adam} optimizer, on a single NVIDIA A100 40 GB GPU.
The values of coefficients $(\alpha, \beta)$ in Equation~\ref{eq:total_loss} were chosen to be $(0.9, 0.1)$ for regression and $(0.7, 0.3)$ for classification.

\paragraph{Baselines.} We use LXMERT jointly trained SR and GQA tasks as a strong baseline for our experiments.
In addition, we also compare performance with existing non-ensemble (single model) methods on the GQA challenge, that directly learn from question-answer pairs without using external program supervision, or additional visual features.
Although NSM~\cite{hudson2019learning} reports a strong performance on the GQA challenge, it uses stronger object detectors and top-50 object features (as opposed to top-36 used by all other baselines), rendering comparison with NSM unfair.

\subsection{Results on Spatial Reasoning}
\input{Latex/tables/comptasks}
\input{Latex/tables/gqa_results}
We begin by evaluating the model on different spatial reasoning tasks, using various weak-supervision training methods. 
Table~\ref{tab:relspa} and~\ref{tab:comptasks} summarize the results for these experiments. 
It can be seen that the LXMERT+SR baseline (trained without supervision from depthmaps) performs poorly for all spatial reasoning tasks.
This conforms with our hypothesis, since depth information is not explicitly captured by the inputs of the current V\&L methods that utilize bounding box information which contains only 2D spatial information.
On average, improvements across SR tasks are correlated with improvements across the GQA task.
In some cases, we observe that the method predicts the correct answer for the spatial relationship questions on the GQA task, even when it fails to correctly predict the bin-classes or object positions in the SR task.
This phenomenon is observed for $18\%$ of the correct GQA predictions.
For example, the model predicts `left' as the GQA answer and a contradictory SR output corresponding to `right'.

\paragraph{Comparison of different SR Tasks.}
Centroid Estimation requires the model to predict the object centroid location in the unit-normalized vector space, whereas the Relative Position Estimation requires the model to determine the pair-wise distance vector between the centroids. 
Both the tasks provide weak-supervision for spatial understanding, but we observe in Table~\ref{tab:comptasks} that bin-classification for the 3D RPE transfers best to the GQA accuracy.

\paragraph{Regression v/s Bin-Classification.}
Similarly, the regression version of the task poses a significant challenge for V\&L models to accurately determine the polarity and the magnitude of distance between the object. 
The range of distances in indoor and outdoor scenes has a large variation, and poses a challenge for the model to exactly predict distances in the regression task.
The classification version of the task appears to be less challenging, with the 3-way 2D relative position estimation achieving significantly high scores (${\sim}90$\%). 
The number of bins (3/15/30) also impacts performance; a larger number of bins implies that the model should possess a fine-grained understanding of distances, which is harder.
We find the optimal number of bins (for both RPE and GQA) is 15.

\paragraph{Comparison of different methods.} The Early Fusion with Image Patches method, which uses both the relative position distance vectors and the pyramidal patch features with the fusion transformer, achieves the best performance across all spatial tasks and the GQA task. 
It can be observed from Table~\ref{tab:relspa} that both of these additional inputs improve performance in 3D RPE.
These performance improvements can be attributed to the direct relation between the distance-vector features and prediction targets.
On the other hand, patch features implicitly possess this spatial relationship information, and utilizing both the features together results in the best performance.
However, even with a direct correlation between the input and output, the model is far from achieving perfect performance on the harder 15/30-way bin-classification or regression tasks, pointing to a scope for further improvements. 

\paragraph{Early v/s Late Fusion.} We can empirically conclude that Early fusion performs better than Late fusion through our experiment results in Table \ref{tab:relspa}. We hypothesize that the Fusion Transformer layers are more efficient than Late Fusion at extracting the spatial relationship information from the projected relative position distance vectors.

\paragraph{Effect of Patch Sizes.}
We study the effect of different image patches' grid sizes, such as $3\times3$, $5\times5$, $7\times7$, and $9\times9$ and several combinations of such sets of patch-features. We observe the best performing feature combination to be the entire image and a set of patches with grids in $3\times3$, $5\times5$ and $7\times7$. Adding smaller patches such as $9\times9$ grid did not lead to an increase in performance. Extracting features from ResNet101 also leads to minor gains ($+0.05\%$).

\subsection{Results on GQA}
Tables~\ref{tab:comp_results} and~\ref{tab:gqa_ood} summarize our results on the GQA and GQA-OOD visual question answering tasks. 
Our best method, LXMERT with Early Fusion and Image Patches, jointly trained with weak-supervision on 15-way bin-classification Relative Position Estimation task improves over the baseline LXMERT, by 1.77\% and 1.3\% respectively on GQA and GQA-OOD, achieving a new state-of-the-art. It performs slightly better than LXMERT (72.9\%) on VQA-v2. 
The most significant improvement is observed on the open-ended questions (2.21\%). 
We can observe that weak-supervision and joint end-to-end training of SR and question answering using the transformer architecture can train systems to be consistent in spatial reasoning tasks and to better generalize in spatial VQA tasks. 

\paragraph{OOD Generalization.}
\input{Latex/tables/gqa_ood}
We also study generalization to distribution shifts for GQA, where the linguistic priors seen during training, undergo a shift at test-time.
We evaluate our best method on the GQA-OOD benchmark and observe that we improve on the most frequent head distribution of answers by $1.7\%$ and also the infrequent out-of-distribution (OOD) tail answer by $0.5\%$. 
This leads us to believe that training on SR tasks with weak-supervision might allows the model to reduce the reliance on spurious linguistic correlations, enabling better generalization abilities.

\begin{figure}[t]
\centering
\includegraphics[width=0.75\linewidth]{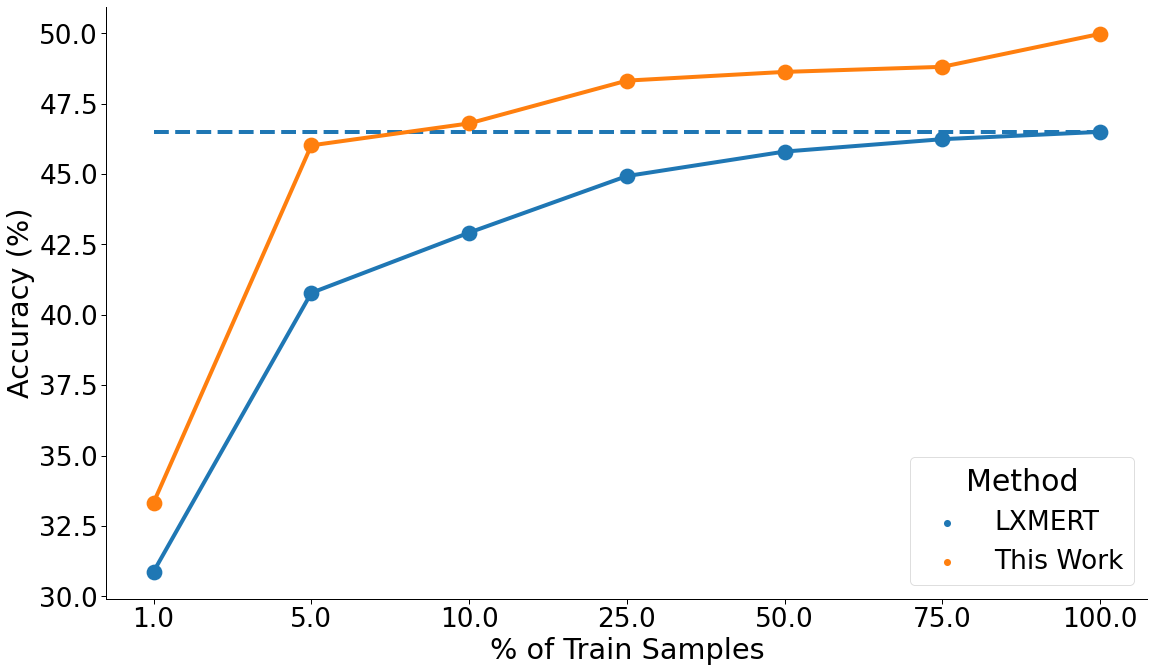}
\caption{Performance of our best method, when trained in the few-shot setting and evaluated on open-ended questions from the GQA-testdev split, compared to LXMERT.}
\label{fig:fewshot}
\end{figure}

\paragraph{Few-Shot Learning.}
We study the effect of the weakly supervised RPE task in the few-shot setting on open-ended questions, with results shown in Figure~\ref{fig:fewshot}. 
We can observe that even with as low as $1\%$ and $5\%$ of samples, joint training with relative position estimation improves over LXMERT trained with same data by $2.5\%$ and $5.5\%$, respectively, and is consistently better than LXMERT at all other fractions.
More importantly, with only $10\%$ of the training dataset our method achieves a performance close to that of the baseline LXMERT trained with the entire ($100\%$) dataset.
Most spatial questions are answered by relative spatial words, such  as ``left'', ``right'', ``up'', ``down'' or object names. 
Object names are learned during the V\&L pre-training tasks, whereas learning about spatial words can be done with few spatial VQA samples and a proper supervision signal that contains spatial information.

\subsection{Error Analysis}
We perform three sets of error analyses to understand the different aspects of the weakly-supervised SR task, the consistency between the relative SR task and the VQA task, and the errors made in the VQA task.

\paragraph{Spatial Reasoning Tasks.} SR-Regression appears to be the most challenging version, as the system needs to reconstruct the relative object distances from the input image to a 3D unit-normalized vector space. 
The classification variant has a higher recall and better polarity, i.e., an object to the ``right'' is classified correctly in the `right' direction regardless of magnitude, i.e.\ the correct distance bin-class, compared to the regression task. 
The majority of errors ($\sim60\%$) are due to the inability to distinguish between close objects. 

\paragraph{Consistency between SR and VQA.} The baseline LXMERT trained only on weak-supervision tasks without patch features or relative position distance vectors predicts $18\%$ of correct predictions with wrong spatial relative positions. This error decreases to $3\%$ for the best method that uses early fusion with image patches, increasing the faithfulness or consistency between the two tasks. We manually analyze 50 inconsistent questions and observe 23 questions contain ambiguity, i.e., multiple objects can be referred by the question and lead to different answers.

\paragraph{Manual Analysis.}
We analyze 100 cases of errors from the GQA test-dev split and broadly categorize them as follows, with percentage of error in parentheses:
\begin{enumerate}[nosep,noitemsep,leftmargin=*]
    \item predictions are \textbf{synonyms or hypernyms} of ground-truth; for example, ``curtains--drapes'', ``cellphone--phone'', ``man--person'', etc.
        \hfill{($8\%$)}
    \item predictions are \textbf{singular/plural} versions of the gold answer, such as, ``curtain-curtains'', ``shelf-shelves''.
        \hfill{($2\%$)}
    \item \textbf{Ambiguous questions} can refer to multiple objects leading to different answers; for example, in an image with two persons having black and brown hair standing in front of a mirror, a question is asked: ``Does the person in front of the mirror have black hair?''.
        \hfill{($5\%$)}
    \item \textbf{Errors in answer annotations.} \hfill{($5\%$)}
    \item \textbf{Wrong predictions}.
    Examples of this include predicting ``right'' when the true answer is ``left'' or the prediction of similar object classes as the answer, such as ``cellphone--remote control'', ``traffic-sign--stop sign''.
    In many cases, the model is able to detect an object, but unable to resolve its relative location with respect to another object; this could be attributed to either spurious linguistic biases or the model's lack of spatial reasoning. \hfill{($80\%$)}
\end{enumerate}
This small-scale study concludes that $20\%$ of the wrong predictions could be mitigated by improved evaluation of subjective, ambiguous, or alternative answers.
Luo~\etal~\cite{luo-etal-2021-just} share this observation and suggest methods for a more robust evaluation of VQA models.

\section{Discussion}
The paradigm of pre-trained models that learn the correspondence between images and text has resulted in improvements across a wide range of V\&L tasks.
Spatial reasoning poses the unique challenge of understanding not only the semantics of the scene, but the physical and geometric properties of the scene.
One stream of work has approached this task from the perspective of sequential instruction-following using program supervision. 
In contrast, our work is the first to jointly model geometric understanding and V\&L in the same training pipeline, via weak supervision from depth estimators.
We show that this increases the faithfulness between spatial reasoning and visual question answering, and improves performance on the GQA dataset in both fully supervised and few-shot settings. 
While in this work, we have used depthmaps as weak supervision, many other concepts from physics-based vision could further come to the aid of V\&L reasoning.
Future work could also consider spatial reasoning in V\&L settings without access to bounding boxes or reliable object detectors (for instance in bad weather and/or low-light settings).
Challenges such as these could potentially reveal the role that geometric and physics-based visual signals could play in robust visual reasoning.

\noindent{}
\textbf{Acknowledgments.}
The authors acknowledge support from NSF grants \#1750082 and \#1816039, DARPA SAIL-ON program \#W911NF2020006, and ONR award \#N00014-20-1-2332.

{\small
\bibliographystyle{ieee_fullname}
\bibliography{egbib}
}

\end{document}

%% file: Latex/01_introduction.tex
\section{Introduction}
\begin{figure}[t]
    \centering
    \includegraphics[width=\linewidth]{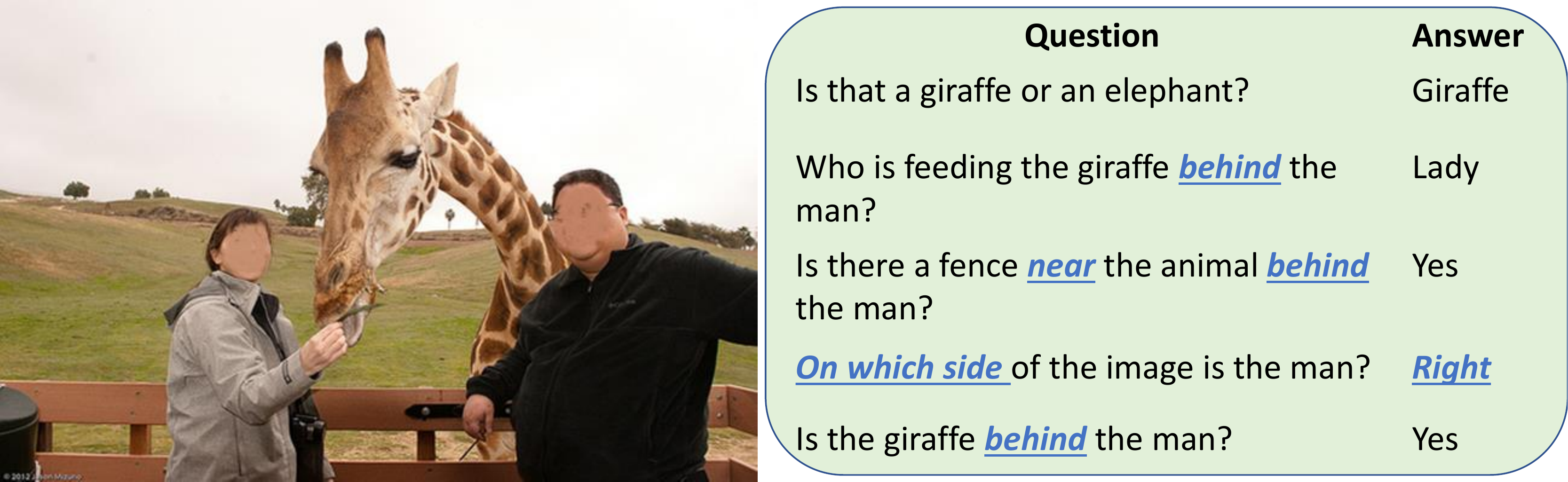}
    \caption{GQA~\cite{hudson2019gqa} requires a compositional understanding of objects, their properties, and spatial locations (underlined).}
    \label{fig:gqa_example}
\end{figure}

``Visual reasoning'' is an umbrella term that is used for visual abilities beyond the perception of appearances (objects and their sizes, shapes, colors, and textures).
In the V\&L domain, tasks such as image-text matching~\cite{suhr2017corpus,suhr2019corpus,vu2018grounded}, visual grounding~\cite{kazemzadeh2014referitgame,yu2016modeling}, visual question answering (VQA)~\cite{goyal2017making,hudson2019gqa}, and commonsense reasoning~\cite{zellers2019recognition} fall under this category.
One such ability is spatial reasoning -- understanding the geometry of the scene and spatial locations of objects in an image.
Visual question answering (such as the GQA challenge shown in Figure~\ref{fig:gqa_example}) is a task that can evaluate this ability via questions that either refer to spatial locations of objects in the image, or questions that require a compositional understanding of spatial relations between objects.

Transformer-based models~\cite{tan2019lxmert,lu2019vilbert,chen2020uniter,gan2020large} have led to significant improvements in multiple V\&L tasks.
However, the underlying training protocol for these models relies on learning correspondences between visual and textual inputs, via pre-training tasks such as image-text matching and cross-modal masked object prediction or feature regression, and then finetuned on specific datasets such as GQA.
As such, these models are not trained to reason about the 3D geometry of the scene, even though the downstream task evaluates spatial understanding.
As a result, V\&L models remain oblivious to the mechanisms of image formation.

\begin{figure}[t]
    \centering
    \includegraphics[width=\linewidth]{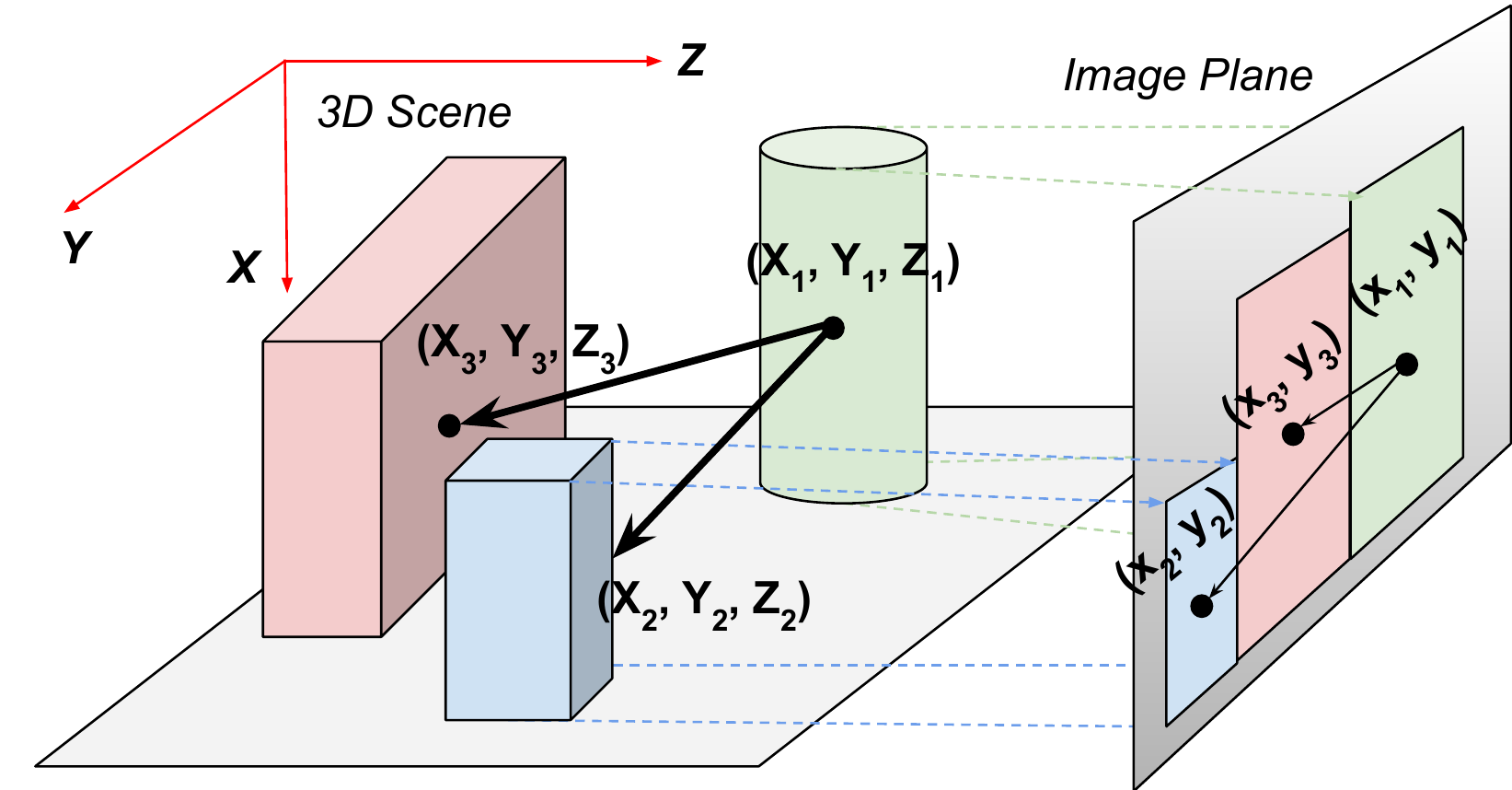}
    \caption{
    When a camera captures an image, points in the 3D scene are projected onto a 2D image plane.
    In VQA, although this projected image is given as input, the questions that require spatial reasoning are inherently about the 3D scene.}
    \label{fig:2d3d}
\end{figure}
\begin{figure}[t]
    \centering
    \includegraphics[height=40mm]{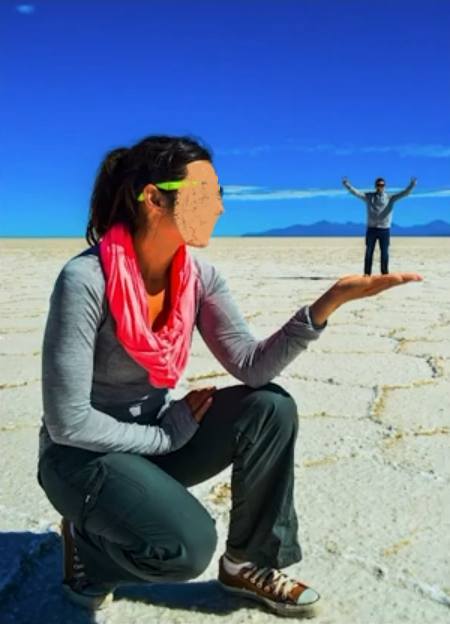}
    \includegraphics[height=40mm]{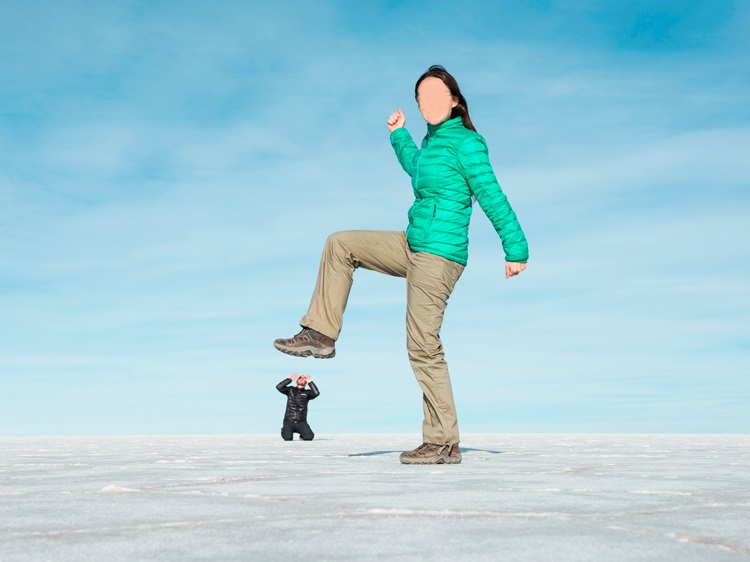}
    \caption{Common optical illusions occur because objects closer to the camera are magnified. This illustrates the need to understand 3D scene geometry to perform spatial reasoning on 2D images.}
    \label{fig:illusion}
\end{figure}

Real-world scenes are 3-dimensional, as illustrated by Figure~\ref{fig:2d3d}, which shows blocks in a scene.
When a camera captures an image of this scene, points on the objects are projected onto the same image plane, i.e., all points get mapped to a single depth value, and the $z$ dimension (depth) is lost.
This mapping depends on lens equations and camera parameters and
leads to optical illusions such as Figure~\ref{fig:illusion}, due to the fact that
the magnification of objects is inversely proportional to the depth and depends on focal lengths~\cite{willson1994modeling,watanabe1996telecentric}.
Since the 3D scene is projected to a 2D image, the faraway person appears smaller, and on top of the woman's palm in the left image, and below the woman's shoe in the right image.
Such relationships between object sizes, depths, camera calibration, and scene geometries make spatial reasoning from images challenging.

If the 3D coordinates of objects $(X_i, Y_i, Z_i)$ are known, it would be trivial to reason about their relative locations, such as the questions in Figure~\ref{fig:gqa_example}.
However, images in V\&L datasets~\cite{hudson2019gqa,goyal2017making} are crowd-sourced and taken from different monocular cameras with unknown and varying camera parameters such as focal length and aperture size, making it difficult to resolve the 3D coordinates (especially the depth) from the image coordinates.
This leads to ambiguities in resolving scene geometry and makes answering questions that require spatial reasoning, a severely ill-posed problem. 

In this paper, we consider the task of visual question answering with emphasis on spatial reasoning (SR).
We investigate if VQA models can resolve spatial relationships between objects in images from the GQA challenge.
Our findings suggest that although models answer some (${\sim}60\%$) of these questions correctly, they cannot faithfully resolve spatial relationships such as relative locations (left, right, front, behind, above, below), or distances between objects.
This opens up a question: 
\begin{quote}
    {\textit{Do VQA models actually understand scene geometry, or do they answer spatial questions based on spurious correlations learned from data?}}
\end{quote}

Towards this end, we design two additional tasks that take 3D geometry into consideration, \emph{object centroid estimation} and \emph{relative position estimation}. 
These tasks are weakly supervised by inferred depth-maps estimated by an off-the-shelf monocular depth-estimation technique~\cite{bhat2020adabins} and bounding box annotations for objects.
For object centroid estimation, the model is trained to predict the centroids of the detected input objects in a unit-normalized 3D vector space. 
On the other hand, for relative position estimation, the model is required to predict the distance vectors between the detected input objects in the same vector space.

Our work can be summarized as follows:
\begin{enumerate}[nosep,noitemsep]
    \item Our approach combined existing training protocosl for transformer-based language models with novel weakly-supervised SR tasks based on the 3D geometry of the scene -- namely, object centroid estimation (OCE) and relative position estimation (RPE).
    \item This approach, significantly improves the correlation between GQA performance and SR tasks.
    \item We show that our approach leads to an improvement of $2.21\%$ on open-ended questions and $1.77\%$ overall, over existing baselines on the GQA challenge.
    \item Our approach also improves the generalization ability to out-of-distribution samples (GQA-OOD~\cite{kervadec2020roses}) and is significantly better than baselines in the few-shot setting achieving state-of-the-art performance with just $10\%$ of labeled GQA samples. 
\end{enumerate}

%% file: Latex/02_related_work.tex
\section{Related Work}
    \textbf{Visual Question Answering}
    is a task at the intersection of vision and language in which systems are expected to answer questions about an image as shown in Figure~\ref{fig:gqa_example}.
    VQA is an active area of research with multiple datasets~\cite{bigham2010vizwiz,antol2015vqa,goyal2017making,hudson2019gqa} that encompass a wide variety of questions, such as questions about the existence of objects and their properties, object counting, questions that require commonsense knowledge~\cite{zellers2019recognition}, external facts or knowledge~\cite{wang2017fvqa,marino2019ok} and spatial reasoning (described below).
    
    \textbf{Spatial Reasoning in VQA}
    has been specifically addressed for synthetic blocks-world images and questions in CLEVR~\cite{johnson2017clevr} and real-world scenes and human-authored questions in GQA~\cite{hudson2019gqa}.
    Both datasets feature questions that require a compositional understanding of spatial relations of objects and their properties. 
    However, the synthetic nature and limited complexity of questions and images in CLEVR make it an easier task;
    models for CLEVR have reached very high ($99.80\%$) test accuracies~\cite{yi2018neural}.
    On the other hand, GQA poses significant challenges owing to the diversity of objects and contexts in real-world scenes and visual ambiguities.
    GQA also brings about linguistic difficulties since questions are crowd-sourced via human annotators.
    For the GQA task, neuro-symbolic methods have been proposed, such as NSM~\cite{hudson2018compositional,hudson2019learning} 
    and TRRNet~\cite{yang2020trrnet} which try to model 
    question-answering as instruction-following by converting questions into symbolic programs.
    
    \textbf{3D scene reconstruction}
    is a fundamental to computer vision and has a long history.
    Depth estimation from multiple observations such as stereo images~\cite{scharstein2002taxonomy}, multiple frames or video~\cite{shroff2012variable,ranftl2016dense}, coded apertures~\cite{zhou2011coded}, variable lighting~\cite{basri2007photometric}, and defocus~\cite{CAVE_0201,tang2017depth} has seen significant progress.
    However monocular (single-image) depth estimation remains a challenging problem, with learning-based methods pushing the envelope~\cite{saxena2005learning,eigen2014depth,li2017two}.
    In this work, we utilize AdaBins~\cite{bhat2020adabins} which uses a transformer-based architecture that adaptively divides depth ranges into variable-sized bins and estimates depth as a linear combination of these depth bins. 
    AdaBins is a state-of-the-art monocular depth estimation model for both outdoor and indoor scenes, and we use it as weak supervision to guide VQA models for spatial reasoning tasks.
    
    \textbf{Weak Supervision in V\&L.}
    Weak supervision is an active area of research in vision tasks such as action/object localization~\cite{song2014learning,zhou2016learning} and semantic segmentation~\cite{khoreva2017simple,zhang2017ppr}.
    While weak supervision \textit{from} V\&L datasets has been used to aid image classification~\cite{ganju2017s,sariyildiz2020icmlm}, the use of weak supervision \textit{for} V\&L and especially for VQA, remains under-explored.
    While existing methodologies have focused on learning cross-modal features from large-scale data, annotations other than objects, questions, and answers have not been extensively used in VQA.
    Kervadec~\etal~\cite{kervadec2019weak} use weak supervision in the form of object-word alignment 
    as a pre-training task, 
    Trott~\etal~\cite{trott2018interpretable} use the counts of objects in an image as weak supervision to guide VQA for counting-based questions, Gokhale~\etal~\cite{gokhale2020vqa} use rules about logical connectives to augment training datasets for yes-no questions, and Zhao~\etal~\cite{zhao2018semantically} use word-embeddings~\cite{mikolov2013distributed} to design an additional weak-supervision objective.
    Weak supervision from captions has also been recently used for visual grounding tasks ~\cite{anne2017localizing,mithun2019weakly,fang2020weak,banerjee-etal-2021-weaqa}.

%% file: Latex/03_relative_spatial_reasoning.tex
\section{Relative Spatial Reasoning}

In V\&L understanding tasks such as image-based VQA, captioning, and visual dialog, systems need to reason about objects present in an image.
Current V\&L systems, such as~\cite{Anderson_2018_CVPR,tan2019lxmert,chen2019uniter,lu2019vilbert} extract FasterRCNN~\cite{ren2015faster} object features to represent the image. 
These systems incorporate positional information by projecting 2D object bounding-box co-ordinates and adding them to the extracted object features.
While V\&L models are pre-trained with tasks such as image-caption matching, masked object prediction, and masked-language modeling, to capture object--word contextual knowledge, none of these tasks explicitly train the system to learn spatial relationships between objects.

In the VQA domain, spatial understanding is evaluated indirectly, by posing questions as shown in Figure~\ref{fig:gqa_example}.
However, this does not objectively capture if the model can infer locations of objects, spatial relations, and distances.
Previous work~\cite{agrawal2018don} has shown that VQA models learn to answer questions by defaulting to spurious linguistic priors between question-answer pairs from the training dataset, which doesn't generalize when the test set undergoes a change in these linguistic priors. 
In a similar vein, our work seeks to disentangle spatial reasoning (SR) from the linguistic priors of the dataset, by introducing two new geometry-based training objectives -- object centroid estimation (OCE) and relative position estimation (RPE).
In this section, we describe these SR tasks.

\subsection{Pre-Processing}
\label{sec:preprocessing}
\paragraph{Pixel Coordinate Normalization.}
We normalize pixel coordinates to the range $[0, 1]$ for both dimensions.
For example, for an image of size $H\times W$, coordinates of a pixel $(x, y)$ are normalized to 
$(\frac{x}{H}, \frac{y}{W})$.

\paragraph{Depth Extraction.}
Although object bounding boxes are available with images in VQA datasets, they lack depth annotations. 
To extract depth-maps from images, we utilize an open-source monocular depth estimation method, AdaBins~\cite{bhat2020adabins}, which is the state-of-the-art on both outdoor~\cite{geiger2013vision} and indoor scene datasets~\cite{silberman2012indoor}.
AdaBins utilizes a transformer that divides an image's depth range into bins whose center value is estimated adaptively per image. 
The final depth values are linear combinations of the bin centers.
As depth values for images lie on vastly different scales for indoor and outdoor images, we normalize depth to the $[0, 1]$ range, using the maximum depth value across all indoor and outdoor images.
We thus obtain depth-values $d(i, j)$ for each pixel $(i, j) , i\in\{1, H\}, j\in\{1, W\}$ in the image.

\paragraph{Representing Objects using Centroids.}
Given the bounding boxes for each object in the image, $[(x_1, y_1), (x_2, y_2)]$ we can compute $(x_c, y_c, z_c)$ coordinates of the object's centroid.
$x_c$ and $y_c$ are calculated as the mean of the top-left corner $(x_1, y_1)$ and bottom-right corner $(x_2, y_2)$ of the bounding box, and $z_c$ is calculated as the mean depth of all points in the bounding box:
\begin{align}
    \small
    x_c &= \frac{x_1 + x_2}{2},
    \quad y_c = \frac{y_1 + y_2}{2} \\
    z_c &= \sum_{\substack{i\in [x_1, x_2],j\in[y_1, y_2].}}d(i, j).
\end{align}
Thus every object $V_k$ in object features can be represented with 3D coordinates of its centroid.
These coordinates act as weak supervision for our spatial reasoning tasks below.

\subsection{Object Centroid Estimation (OCE)}
Our first spatial reasoning task trains models to predict centroids of each object in the image.

In \textbf{2D OCE}, we model the task as prediction of the 2D centroid co-ordinates $(x_c, y_c)$ of the input objects. 
Let $V$ denote the features of the input image and let $Q$ be the textual input.
Then the 2D estimation task requires the system to predict the centroid co-ordinates, 
$(x_{c_k},y_{c_k})$, for all objects $k\in \{1\dots N\}$ present in object-features $V$.

In \textbf{3D OCE}, we also predict the depth co-ordinate of the object. 
Hence the task requires the system to predict the 3D centroid co-ordinates, $(x_{c_k},y_{c_k}, z_{c_k})$, for all objects $k\in \{1\dots N\}$ present in object-features $V$.

\subsection{Relative Position Estimation (RPE)}
\label{relpos}
The model is trained to predict the distance vector between each pair of distinct objects in the projected unit-normalized vector space. 
These distance vectors real-valued vectors $\in \mathbb{R}^{3}_{[-1, 1]}$.
Therefore, for a pair of centroids $(x_1, y_1, z_1)$ and $(x_2, y_2, z_2)$ for two distinct objects, given $V$ and $Q$, the model is trained to predict the vector $[x_1-x_2,y_1-y_2,z_1-z_2]$.
RPE is not symmetric and for any two distinct points $A, B$, $\texttt{dist}(A,B)  = -\texttt{dist}(B,A)$. 

\paragraph{Regression vs. Bin Classification.}
In both tasks above, predictions are real-valued vectors. 
Hence, we evaluate two variants of these tasks: (1) a regression task, where models predict real-valued vectors in $\mathbb{R}^{3}_{[-1, 1]}$, and (2) bin classification, for which we divide the range of real values across all three dimensions into $C$ log-scale bins.
Bin-width for the $c^{th}$ bin is given by (with hyper-parameter $\lambda=1.5$):
\begin{equation}
    b_c = \frac{1}{\lambda^{C - |c-\frac{C}{2}| + 1}} - \frac{1}{\lambda^{C - |c-   \frac{C}{2}| +2}} ~~ \forall c \in \{0..C\text{-}1\}.
\end{equation}
Log-scale bins lead to a higher resolution (more bins) for closer distances and lower resolution (fewer bins) for farther distances, giving us fine-grained classes for close objects.
Models are trained to predict the bin classes as outputs for all 3 dimensions, given a pair of objects.
We evaluate different values for the number of bins: $C \in \{3, 7, 15, 30\}$, to study the extent of V\&L model's ability to differentiate at a higher resolution of spatial distances. 
For example, the simplest form of bin classification is a three-class classification task with bin-intervals $[-1,0)$, $[0]$ , $(0,1]$.

%% file: Latex/04_method.tex
\section{Method}

\begin{figure*}[h]
    \centering
    \includegraphics[width=\textwidth]{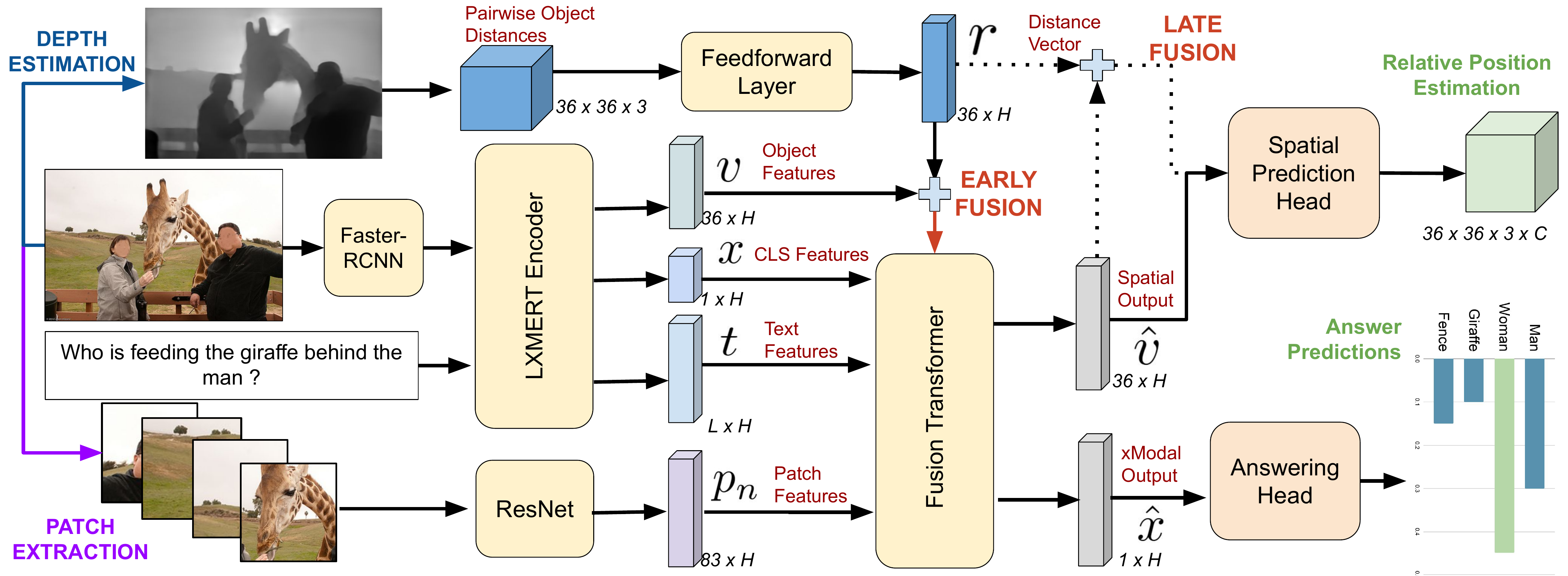}
    \caption{
        Overall architecture for our approach shows conventional modules for object feature extraction, cross-modal encoding, and answering head, with our novel weak supervision from depthmaps, patch extraction, fusion mechanisms, and spatial prediction head.}
    \label{fig:arch}
\end{figure*}

We adopt LXMERT~\cite{tan2019lxmert}, a state-of-the-art vision and language model, as the backbone for our experiments.
LXMERT and other popular transformer-based V\&L models methods~\cite{lu2019vilbert,chen2019uniter}, are pre-trained on a combination of multiple VQA and image captioning datasets such as Conceptual Captions~\cite{sharma2018conceptual}, SBU Captions~\cite{ordonez2011im2text}, Visual Genome~\cite{krishna2017visual}, and MSCOCO~\cite{lin2014microsoft}. 
These models use object features of the top 36 objects extracted by the FasterRCNN object detector~\cite{ren2015faster} as visual representations for input images.
A transformer encoder takes these object features along with textual features as inputs, and outputs cross-modal \texttt{[CLS]} tokens.
The model is pre-trained by optimizing for masked language modeling, image-text matching, masked-object prediction and image-question answering.

\subsection{Weak Supervision for SR}
\label{weaksup}
Let $v\in\mathbb{R}^{36\times H}$ be the visual features, $x\in\mathbb{R}^{1\times H}$ be the cross-modal features, and $t\in\mathbb{R}^{L\times H}$ be the text features, produced by the cross-modality attention layer of the LXMERT encoder.
Here $H$ is the hidden dimension, and $L$ is the number of tokens.
These outputs are used for fine-tuning the model for two tasks:
VQA using $x$ as input, 
and the spatial reasoning tasks using $v$ as input. 
Let $D$ be the number of coordinate dimensions (2 or 3) that we use in spatial reasoning.
For the SR-regression task, we use a two-layer feed-forward network $f_{reg}$ to project $v$ to a real-valued vector with dimensions $36\times D$, and compute the loss using mean-squared error (MSE) with respect to the ground-truth object coordinates $y_{reg}$ .
\begin{equation}
    \small 
    \mathcal{L}_{SR\text{-}reg} = \mathcal{L}_{MSE} (f_{reg}(v), y_{reg}).
\end{equation}

For the bin-classification task, we train a two-layer feed-forward network $f_{bin}$ to predict $36\times C\times D$ bin classes for each object along each dimension, where $C$ is the number of classes, trained using cross-entropy loss:
\begin{equation}
    \small 
    \mathcal{L}_{SR\text{-}bin} = \mathcal{L}_{CE} (f_{bin}(V), y_{bin}), 
\end{equation}
where $y_{bin}$ are the ground-truth object location bins.

\noindent The total loss is given by:
\begin{equation}
    \small
    \mathcal{L} =  \alpha\cdot\mathcal{L}_{VQA} + \beta\cdot\mathcal{L}_{SR}, \quad where~\alpha,\beta \in (0, 1].
    \label{eq:total_loss}
\end{equation}
$y_{reg}$ and $y_{bin}$ are obtained from the object centroids computed during preprocessing (Sec.~\ref{sec:preprocessing}) from depth estimation networks and object bounding boxes.
Since the real 3D coordinates of objects in the scene are unknown, these $y_{reg}$ and $y_{bin}$ act as proxies and therefore can weakly supervise our spatial reasoning tasks.

\subsection{Spatial Pyramid Patches}
As LXMERT only takes as input the distinct object and the 2D bounding box features, it inherently lacks the depth information required for 3D spatial reasoning task. 
This is confirmed by our evaluation on the 2D and 3D spatial reasoning tasks, where the model has strong performance in 2D tasks, but lacks on 3D tasks, as shown in Table~\ref{tab:relspa}. 
In order to incorporate spatial features from the original image to capture relative object locations as well as depth information, we propose the use of \textit{spatial pyramid patch features}~\cite{banerjee-etal-2021-weaqa} to represent the given image into a sequence of features at different scales. 
The image $I$ is divided into a set of patches: $p_n = \{ I_{i_1},\dots, I_{i_n}\}$, each $I_{i_j}$ being a $i_j\times i_j$ grid of patches, and ResNet features are extracted for each patch.
Larger patches encode global object relationships, while smaller patches contain local relationships.

\subsection{Fusion Transformer}
In order to combine the spatial pyramid patch features and features extracted from LXMERT, we propose a fusion transformer with $e$-layers of transformer encoders, containing self-attention, a residual connection and layer normalization after each sub-layer. 
We concatenate the $p_n$ patch features with $v$ visual, $x$ cross-modal and $t$ textual hidden vector output representations from LXMERT, to create the fused vector $h$, which is fed into the fusion transformer. 
Let $M$ be the length of the sequence after concatenating all hidden vectors, then for any hidden vector $m$ in the sequence:
\begin{align}
    h^0 &= [X,V,T,P_n]. \nonumber \\
    \hat{h}^{e}_m &= \texttt{Self-Att}(h^{e-1}_{m},[h^{e-1}_1,\dots,h^{e-1}_{M}]); ~\forall e.
\end{align}
The output of fusion transformer $\hat{h}^e = [\hat{x}, \hat{v}, \hat{t}, \hat{p}_n]$ is then separated into its components, of which, 
$\hat{x}$, $\hat{v}$ are used as inputs for VQA and SR task, on the same lines as Section~\ref{weaksup}. 

\subsection{Relative Position Vectors as Inputs}
The final set of features that we utilize are the pair-wise relative distance vectors between objects as described in section~\ref{relpos}. 
In this case, the pairwise distances are used as inputs, in addition to visual, textual, cross-modal and patch features, and the model is trained to reconstruct the pairwise distances.
This makes our model an auto-encoder for the regression task.
For each input visual object feature $v_k$, we create a relative position feature $r_k$ using the pair-wise distance vectors projected from the input dimensions of $36\times 3$ to $36\times H$ using a feed-forward layer, where $H$ is the size of the hidden vector representations. 
We evaluate two-modes of fusion of these features. 
In \textbf{Early Fusion}, $r_k$ is added to $v_k$ the output of the LXMERT encoder. 
In \textbf{Late Fusion}, $r_k$ is added to $\hat{v}_k$ the output of the fusion transformer. 
Figure \ref{fig:arch} shows the architecture for the final model that utilizes both the patch features and relative positions as input. 

%% file: Latex/tables/spatial_task_2d_3d.tex
\begin{table*}[t]
    \centering
    \small
    \resizebox{\linewidth}{!}{
    \begin{tabular}{@{}l ccccc c ccccc@{}}
        \toprule 
        \multirow{2}{*}{\textbf{Model}} & 
        
        \multirow{2}{*}{\textbf{GQA-Val}$\uparrow$} & \multirow{2}{*}{\textbf{2D-Reg}$\downarrow$}& \multicolumn{3}{c}{2D Bin Classification}
        && \multirow{2}{*}{\textbf{GQA-Val}$\uparrow$} & \multirow{2}{*}{\textbf{3D-Reg}$\downarrow$} & \multicolumn{3}{c}{3D Bin Classification}
        \\
        \cmidrule{4-6} \cmidrule{10-12}
        &&  & \textbf{2D-3w}$\uparrow$ & \textbf{2D-15w}$\uparrow$ & \textbf{2D-30w}$\uparrow$ &&  & & \textbf{3D-3w}$\uparrow$ & \textbf{3D-15w}$\uparrow$ & \textbf{3D-30w}$\uparrow$ \\
        \midrule 
        LXMERT~+~SR                          & 59.85  & 0.64 & 88.20 & 76.75 & 55.12 && 60.05 & 0.44 & 55.66 & 52.80 &  48.15 \\
        \quad~+~Late Fusion            & 59.90  & 0.47 & 92.60 & 81.24 & 60.42 && 60.18 & 0.29 & 71.20 & 69.45 &  52.84 \\
        \quad~+~Early Fusion           & 60.10  & 0.36 & 96.40 & 82.48 & 64.85 && 61.32 & 0.24 & 78.67 & 74.20 &  54.73 \\
        \quad~+~Patches                & 60.52  & 0.41 & 89.60 & 79.56 & 59.40 && 60.64 & 0.28 & 73.21 & 71.74 &  50.94  \\
        \quad~+~Late Fusion~+~Patches  & 60.80  & 0.33 & 95.20 & 82.10 & 67.38 && 61.80 & 0.21 & 85.35 & 79.60 &  65.45  \\
        \quad~+~Early Fusion~+~Patches & \textbf{60.95}  & \textbf{0.29} & \textbf{97.40} & \textbf{84.60} & \textbf{71.46} && \textbf{62.32} & \textbf{0.17} & \textbf{89.58} & \textbf{81.47} &  \textbf{68.20} \\
        \bottomrule
    \end{tabular}
    }
    \caption{Results for the LXMERT model trained for the spatial reasoning task (LXMERT~+~SR), on 2D and 3D Relative Position Estimation (RPE), for regression as well as C-way bin classification tasks.
    A comparison with the same model weakly supervised with additional features (image patches) and weak supervision with relative position vectors extracted from depth-maps is shown.
    GQA-Val scores are for the best performing weak-supervision task, which are 2D-15w and 3D-15w respectively. 
    Regression scores are in terms of mean-squared error, and classification scores are percentage accuracy. \textit{15w: 15-way bin-classification.}
    }
    \label{tab:relspa}
\end{table*}

%% file: Latex/tables/comptasks.tex
\begin{table}[t]
    \centering
    \small
    \begin{tabular*}{\linewidth}{@{\extracolsep{\fill}}lc@{}}
        \toprule
        \textbf{Model} & \textbf{GQA-Val$\uparrow$} \\ 
        \midrule 
        LXMERT~+~SR & 59.40 \\
        \multicolumn{2}{c}{\makebox[\linewidth]{\dashrule}}\\[-\jot]
        \quad+ 2D OCE (Regression) & 57.33 \\
        \quad+ 3D OCE (Regression) & 58.28 \\
        \quad+ 2D RPE (Regression) & 59.85 \\
        \quad+ 3D RPE (Regression) & 59.54 \\
        \multicolumn{2}{@{}c@{}}{\makebox[\linewidth]{\dashrule}}\\[-\jot]
        \quad+ 2D OCE (15-bin Classification) & 58.64 \\
        \quad+ 3D OCE (15-bin Classification) & 59.90 \\
        \quad+ 2D RPE (15-bin Classification) & 60.95 \\
        \quad+ 3D RPE (15-bin Classification) & \textbf{62.32} \\ 
        \bottomrule
    \end{tabular*}
    \caption{Comparison of different weakly supervised spatial reasoning tasks on the GQA validation split. }    
    \label{tab:comptasks}
\end{table}

%% file: Latex/tables/gqa_results.tex
\begin{table*}[t]
    \centering
    \small
    \begin{tabular*}{\linewidth}{@{\extracolsep{\fill}}lccccccc@{}}
        \toprule
        \textbf{Model} & \textbf{Accuracy}$\uparrow$ & \textbf{Binary}$\uparrow$ & \textbf{Open}$\uparrow$ & \textbf{Consistency}$\uparrow$ & \textbf{Validity}$\uparrow$ & \textbf{Plausibility}$\uparrow$ & \textbf{Distribution}$\downarrow$ \\ \midrule
        Human~\cite{hudson2019gqa}            & 89.30 & 91.20 & 87.40 & 98.40 & 98.90 & 97.20 & -- \\
        Global Prior~\cite{hudson2019gqa}     & 28.90 & 42.94 & 16.62 & 51.69 & 88.86 & 74.81 & 93.08 \\
        Local Prior~\cite{hudson2019gqa}     & 31.24 & 47.90 & 16.66 & 54.04 & 84.33 & 84.31 & 13.98 \\
        BottomUp~\cite{Anderson_2018_CVPR}        & 49.74 & 66.64 & 34.83 & 78.71 & 96.18 & 84.57 & 5.98 \\
        MAC~\cite{hudson2018compositional}             & 54.06 & 71.23 & 38.91 & 81.59 & 96.16 & 84.48 & 5.34 \\
        GRN~\cite{hudson2019learning}             & 59.37 & 77.53 & 43.35 & 88.63 & 96.18 & 84.71 & 6.06 \\
        Dream~\cite{hudson2019learning}           & 59.72 & 77.84 & 43.72 & 91.71 & 96.38 & \textbf{85.48} & 8.40 \\
        LXMERT~\cite{tan2019lxmert}
        & 60.34 & 77.76 & 44.97 & 92.84 & 96.30 & 85.19 & 8.31 \\
        This Work       & \textbf{62.11} & \textbf{78.20} & \textbf{47.18} & \textbf{93.13} &  \textbf{96.92}  & 85.27 &\textbf{ 1.10} \\ 
        \bottomrule
    \end{tabular*}
    \caption{Comparitive evaluation of our model with respect to existing baselines, on the GQA test-standard set, along all evaluation metrics.
    }
    \label{tab:comp_results}
\end{table*}


%% file: Latex/tables/gqa_ood.tex
\begin{table}[t]
    \centering
    \small
    \resizebox{\linewidth}{!}{
    \begin{tabular}{@{}lccccc@{}}
        \toprule
        \textbf{Model} & \textbf{Uses Image} & \textbf{Acc-All}$\uparrow$ & \textbf{Acc-Tail}$\uparrow$ & \textbf{Acc-Head}$\uparrow$  \\ \midrule
        Question Prior~\cite{kervadec2020roses} & No & 21.6 & 17.8 & 24.1  \\
        LSTM~\cite{antol2015vqa} & No & 30.7 & 24.0 & 34.8  \\
        BottomUp~\cite{Anderson_2018_CVPR} & Yes & 46.4 & 42.1 & 49.1  \\
        MCAN~\cite{yu2019deep} & Yes & 50.8 & 46.5 & 53.4  \\
        BAN4~\cite{kim2018bilinear} & Yes & 50.2 & 47.2 & 51.9  \\
        MMN~\cite{chen2021meta} & Yes & 52.7 & 48.0 & 55.5  \\
        LXMERT~\cite{tan2019lxmert} & Yes & 54.6 & 49.8 & 57.7 \\
        This Work & Yes & \textbf{55.9}  & \textbf{50.3} & \textbf{59.4} \\
        \bottomrule
    \end{tabular}
    }   
    \caption{Comparison of several VQA methods on the GQA-OOD test-dev splits. Acc-tail: OOD settings, Acc-head: accuracy on most probable answers (given context), scores in \%.}
    \label{tab:gqa_ood}
\end{table}

%% file: egpaper_for_review.bbl
\begin{thebibliography}{10}\itemsep=-1pt

\bibitem{agrawal2018don}
Aishwarya Agrawal, Dhruv Batra, Devi Parikh, and Aniruddha Kembhavi.
\newblock Don't just assume; look and answer: Overcoming priors for visual
  question answering.
\newblock In {\em CVPR}, 2018.

\bibitem{Anderson_2018_CVPR}
Peter Anderson, Xiaodong He, Chris Buehler, Damien Teney, Mark Johnson, Stephen
  Gould, and Lei Zhang.
\newblock Bottom-up and top-down attention for image captioning and visual
  question answering.
\newblock In {\em CVPR}, June 2018.

\bibitem{antol2015vqa}
Stanislaw Antol, Aishwarya Agrawal, Jiasen Lu, Margaret Mitchell, Dhruv Batra,
  C~Lawrence Zitnick, and Devi Parikh.
\newblock Vqa: Visual question answering.
\newblock In {\em Proceedings of the IEEE international conference on computer
  vision}, pages 2425--2433, 2015.

\bibitem{banerjee-etal-2021-weaqa}
Pratyay Banerjee, Tejas Gokhale, Yezhou Yang, and Chitta Baral.
\newblock {W}ea{QA}: Weak supervision via captions for visual question
  answering.
\newblock In {\em Findings of the Association for Computational Linguistics:
  ACL-IJCNLP 2021}, pages 3420--3435, Online, Aug. 2021. Association for
  Computational Linguistics.

\bibitem{basri2007photometric}
Ronen Basri, David Jacobs, and Ira Kemelmacher.
\newblock Photometric stereo with general, unknown lighting.
\newblock {\em International Journal of computer vision}, 72(3):239--257, 2007.

\bibitem{bhat2020adabins}
Shariq~Farooq Bhat, Ibraheem Alhashim, and Peter Wonka.
\newblock Adabins: Depth estimation using adaptive bins.
\newblock {\em arXiv preprint arXiv:2011.14141}, 2020.

\bibitem{bigham2010vizwiz}
Jeffrey~P Bigham, Chandrika Jayant, Hanjie Ji, Greg Little, Andrew Miller,
  Robert~C Miller, Robin Miller, Aubrey Tatarowicz, Brandyn White, Samual
  White, et~al.
\newblock Vizwiz: nearly real-time answers to visual questions.
\newblock In {\em Proceedings of the 23nd annual ACM symposium on User
  interface software and technology}, pages 333--342, 2010.

\bibitem{chen2021meta}
Wenhu Chen, Zhe Gan, Linjie Li, Yu Cheng, William Wang, and Jingjing Liu.
\newblock Meta module network for compositional visual reasoning.
\newblock In {\em Proceedings of the IEEE/CVF Winter Conference on Applications
  of Computer Vision}, pages 655--664, 2021.

\bibitem{chen2019uniter}
Yen-Chun Chen, Linjie Li, Licheng Yu, Ahmed~El Kholy, Faisal Ahmed, Zhe Gan, Yu
  Cheng, and Jingjing Liu.
\newblock Uniter: Learning universal image-text representations.
\newblock {\em arXiv preprint arXiv:1909.11740}, 2019.

\bibitem{chen2020uniter}
Yen-Chun Chen, Linjie Li, Licheng Yu, Ahmed~El Kholy, Faisal Ahmed, Zhe Gan, Yu
  Cheng, and Jingjing Liu.
\newblock Uniter: Universal image-text representation learning.
\newblock In {\em ECCV}, 2020.

\bibitem{eigen2014depth}
David Eigen, Christian Puhrsch, and Rob Fergus.
\newblock Depth map prediction from a single image using a multi-scale deep
  network.
\newblock {\em Advances in Neural Information Processing Systems},
  27:2366--2374, 2014.

\bibitem{fang2020weak}
Zhiyuan Fang, Shu Kong, Zhe Wang, Charless Fowlkes, and Yezhou Yang.
\newblock Weak supervision and referring attention for temporal-textual
  association learning.
\newblock {\em arXiv preprint arXiv:2006.11747}, 2020.

\bibitem{gan2020large}
Zhe Gan, Yen-Chun Chen, Linjie Li, Chen Zhu, Yu Cheng, and Jingjing Liu.
\newblock Large-scale adversarial training for vision-and-language
  representation learning.
\newblock In {\em NeurIPS}, 2020.

\bibitem{ganju2017s}
Siddha Ganju, Olga Russakovsky, and Abhinav Gupta.
\newblock What's in a question: Using visual questions as a form of
  supervision.
\newblock In {\em Proceedings of the IEEE Conference on Computer Vision and
  Pattern Recognition}, pages 241--250, 2017.

\bibitem{geiger2013vision}
Andreas Geiger, Philip Lenz, Christoph Stiller, and Raquel Urtasun.
\newblock Vision meets robotics: The kitti dataset.
\newblock {\em The International Journal of Robotics Research},
  32(11):1231--1237, 2013.

\bibitem{gokhale2020vqa}
Tejas Gokhale, Pratyay Banerjee, Chitta Baral, and Yezhou Yang.
\newblock Vqa-lol: Visual question answering under the lens of logic.
\newblock In {\em European Conference on Computer Vision (ECCV)}, 2020.

\bibitem{goyal2017making}
Yash Goyal, Tejas Khot, Douglas Summers-Stay, Dhruv Batra, and Devi Parikh.
\newblock Making the v in vqa matter: Elevating the role of image understanding
  in visual question answering.
\newblock In {\em Proceedings of the IEEE Conference on Computer Vision and
  Pattern Recognition}, pages 6904--6913, 2017.

\bibitem{he2016deep}
Kaiming He, Xiangyu Zhang, Shaoqing Ren, and Jian Sun.
\newblock Deep residual learning for image recognition.
\newblock In {\em CVPR}, pages 770--778, 2016.

\bibitem{anne2017localizing}
Lisa-Anne Hendricks, Oliver Wang, Eli Shechtman, Josef Sivic, Trevor Darrell,
  and Bryan Russell.
\newblock Localizing moments in video with natural language.
\newblock In {\em Proceedings of the IEEE international conference on computer
  vision}, pages 5803--5812, 2017.

\bibitem{hudson2018compositional}
Drew~A Hudson and Christopher~D Manning.
\newblock Compositional attention networks for machine reasoning.
\newblock In {\em International Conference on Learning Representations}, 2018.

\bibitem{hudson2019gqa}
Drew~A Hudson and Christopher~D Manning.
\newblock Gqa: A new dataset for real-world visual reasoning and compositional
  question answering.
\newblock In {\em Proceedings of the IEEE/CVF Conference on Computer Vision and
  Pattern Recognition}, pages 6700--6709, 2019.

\bibitem{hudson2019learning}
Drew~A Hudson and Christopher~D Manning.
\newblock Learning by abstraction: The neural state machine.
\newblock {\em arXiv preprint arXiv:1907.03950}, 2019.

\bibitem{johnson2017clevr}
Justin Johnson, Bharath Hariharan, Laurens Van Der~Maaten, Li Fei-Fei, C
  Lawrence~Zitnick, and Ross Girshick.
\newblock Clevr: A diagnostic dataset for compositional language and elementary
  visual reasoning.
\newblock In {\em Proceedings of the IEEE Conference on Computer Vision and
  Pattern Recognition}, pages 2901--2910, 2017.

\bibitem{kazemzadeh2014referitgame}
Sahar Kazemzadeh, Vicente Ordonez, Mark Matten, and Tamara Berg.
\newblock Referitgame: Referring to objects in photographs of natural scenes.
\newblock In {\em Proceedings of the 2014 conference on empirical methods in
  natural language processing (EMNLP)}, pages 787--798, 2014.

\bibitem{kervadec2019weak}
Corentin Kervadec, Grigory Antipov, Moez Baccouche, and Christian Wolf.
\newblock Weak supervision helps emergence of word-object alignment and
  improves vision-language tasks.
\newblock {\em arXiv preprint arXiv:1912.03063}, 2019.

\bibitem{kervadec2020roses}
Corentin Kervadec, Grigory Antipov, Moez Baccouche, and Christian Wolf.
\newblock Roses are red, violets are blue... but should vqa expect them to?
\newblock {\em CVPR}, 2021.

\bibitem{khoreva2017simple}
Anna Khoreva, Rodrigo Benenson, Jan Hosang, Matthias Hein, and Bernt Schiele.
\newblock Simple does it: Weakly supervised instance and semantic segmentation.
\newblock In {\em Proceedings of the IEEE conference on computer vision and
  pattern recognition}, pages 876--885, 2017.

\bibitem{kim2018bilinear}
Jin-Hwa Kim, Jaehyun Jun, and Byoung-Tak Zhang.
\newblock Bilinear attention networks.
\newblock In {\em Advances in Neural Information Processing Systems}, pages
  1564--1574, 2018.

\bibitem{kingma2014adam}
Diederik~P Kingma and Jimmy Ba.
\newblock Adam: A method for stochastic optimization.
\newblock {\em arXiv preprint arXiv:1412.6980}, 2014.

\bibitem{krishna2017visual}
Ranjay Krishna, Yuke Zhu, Oliver Groth, Justin Johnson, Kenji Hata, Joshua
  Kravitz, Stephanie Chen, Yannis Kalantidis, Li-Jia Li, David~A Shamma, et~al.
\newblock Visual genome: Connecting language and vision using crowdsourced
  dense image annotations.
\newblock {\em International journal of computer vision}, 123(1):32--73, 2017.

\bibitem{li2017two}
Jun Li, Reinhard Klein, and Angela Yao.
\newblock A two-streamed network for estimating fine-scaled depth maps from
  single rgb images.
\newblock In {\em Proceedings of the IEEE International Conference on Computer
  Vision}, pages 3372--3380, 2017.

\bibitem{lin2014microsoft}
Tsung-Yi Lin, Michael Maire, Serge Belongie, James Hays, Pietro Perona, Deva
  Ramanan, Piotr Doll{\'a}r, and C~Lawrence Zitnick.
\newblock Microsoft coco: Common objects in context.
\newblock In {\em European conference on computer vision}, pages 740--755.
  Springer, 2014.

\bibitem{lu2019vilbert}
Jiasen Lu, Dhruv Batra, Devi Parikh, and Stefan Lee.
\newblock Vilbert: Pretraining task-agnostic visiolinguistic representations
  for vision-and-language tasks.
\newblock In {\em Advances in Neural Information Processing Systems}, pages
  13--23, 2019.

\bibitem{luo-etal-2021-just}
Man Luo, Shailaja~Keyur Sampat, Riley Tallman, Yankai Zeng, Manuha Vancha,
  Akarshan Sajja, and Chitta Baral.
\newblock {`}just because you are right, doesn{'}t mean {I} am wrong{'}:
  Overcoming a bottleneck in development and evaluation of open-ended {VQA}
  tasks.
\newblock In {\em Proceedings of the 16th Conference of the European Chapter of
  the Association for Computational Linguistics: Main Volume}, pages
  2766--2771, Online, Apr. 2021. Association for Computational Linguistics.

\bibitem{marino2019ok}
Kenneth Marino, Mohammad Rastegari, Ali Farhadi, and Roozbeh Mottaghi.
\newblock Ok-vqa: A visual question answering benchmark requiring external
  knowledge.
\newblock In {\em Proceedings of the IEEE/CVF Conference on Computer Vision and
  Pattern Recognition}, pages 3195--3204, 2019.

\bibitem{mikolov2013distributed}
Tomas Mikolov, Ilya Sutskever, Kai Chen, Greg Corrado, and Jeffrey Dean.
\newblock Distributed representations of words and phrases and their
  compositionality.
\newblock In {\em Neural Information Processing Systems}, NIPS'13, page
  3111–3119, Red Hook, NY, USA, 2013. Curran Associates Inc.

\bibitem{mithun2019weakly}
Niluthpol~Chowdhury Mithun, Sujoy Paul, and Amit~K Roy-Chowdhury.
\newblock Weakly supervised video moment retrieval from text queries.
\newblock In {\em Proceedings of the IEEE/CVF Conference on Computer Vision and
  Pattern Recognition}, pages 11592--11601, 2019.

\bibitem{ordonez2011im2text}
Vicente Ordonez, Girish Kulkarni, and Tamara~L Berg.
\newblock Im2text: Describing images using 1 million captioned photographs.
\newblock In {\em Advances in neural information processing systems}, 2011.

\bibitem{ranftl2016dense}
Rene Ranftl, Vibhav Vineet, Qifeng Chen, and Vladlen Koltun.
\newblock Dense monocular depth estimation in complex dynamic scenes.
\newblock In {\em Proceedings of the IEEE conference on computer vision and
  pattern recognition}, pages 4058--4066, 2016.

\bibitem{ren2015faster}
Shaoqing Ren, Kaiming He, Ross Girshick, and Jian Sun.
\newblock Faster r-cnn: Towards real-time object detection with region proposal
  networks.
\newblock In {\em Advances in neural information processing systems}, pages
  91--99, 2015.

\bibitem{russakovsky2015imagenet}
Olga Russakovsky, Jia Deng, Hao Su, Jonathan Krause, Sanjeev Satheesh, Sean Ma,
  Zhiheng Huang, Andrej Karpathy, Aditya Khosla, Michael Bernstein, et~al.
\newblock Imagenet large scale visual recognition challenge.
\newblock {\em International journal of computer vision}, 115(3), 2015.

\bibitem{sariyildiz2020icmlm}
Mert~Bulent Sariyildiz, Julien Perez, and Diane Larlus.
\newblock Learning visual representations with caption annotations.
\newblock In {\em European Conference on Computer Vision (ECCV)}, 2020.

\bibitem{saxena2005learning}
Ashutosh Saxena, Sung~H Chung, Andrew~Y Ng, et~al.
\newblock Learning depth from single monocular images.
\newblock In {\em NIPS}, volume~18, pages 1--8, 2005.

\bibitem{scharstein2002taxonomy}
Daniel Scharstein and Richard Szeliski.
\newblock A taxonomy and evaluation of dense two-frame stereo correspondence
  algorithms.
\newblock {\em International journal of computer vision}, 47(1):7--42, 2002.

\bibitem{sharma2018conceptual}
Piyush Sharma, Nan Ding, Sebastian Goodman, and Radu Soricut.
\newblock Conceptual captions: A cleaned, hypernymed, image alt-text dataset
  for automatic image captioning.
\newblock In {\em Proceedings of the 56th Annual Meeting of the ACL}, pages
  2556--2565, 2018.

\bibitem{shroff2012variable}
Nitesh Shroff, Ashok Veeraraghavan, Yuichi Taguchi, Oncel Tuzel, Amit Agrawal,
  and Rama Chellappa.
\newblock Variable focus video: Reconstructing depth and video for dynamic
  scenes.
\newblock In {\em 2012 IEEE International Conference on Computational
  Photography (ICCP)}, pages 1--9. IEEE, 2012.

\bibitem{silberman2012indoor}
Nathan Silberman, Derek Hoiem, Pushmeet Kohli, and Rob Fergus.
\newblock Indoor segmentation and support inference from rgbd images.
\newblock In {\em European conference on computer vision}, pages 746--760.
  Springer, 2012.

\bibitem{song2014learning}
Hyun~Oh Song, Ross Girshick, Stefanie Jegelka, Julien Mairal, Zaid Harchaoui,
  and Trevor Darrell.
\newblock On learning to localize objects with minimal supervision.
\newblock In {\em International Conference on Machine Learning}, pages
  1611--1619. PMLR, 2014.

\bibitem{suhr2017corpus}
Alane Suhr, Mike Lewis, James Yeh, and Yoav Artzi.
\newblock A corpus of natural language for visual reasoning.
\newblock In {\em Proceedings of the 55th Annual Meeting of the Association for
  Computational Linguistics (Volume 2: Short Papers)}, pages 217--223, 2017.

\bibitem{suhr2019corpus}
Alane Suhr, Stephanie Zhou, Ally Zhang, Iris Zhang, Huajun Bai, and Yoav Artzi.
\newblock A corpus for reasoning about natural language grounded in
  photographs.
\newblock In {\em Proceedings of the Annual Meeting of the Association for
  Computational Linguistics}, 2019.

\bibitem{tan2019lxmert}
Hao Tan and Mohit Bansal.
\newblock Lxmert: Learning cross-modality encoder representations from
  transformers.
\newblock In {\em EMNLP 2019}, 2019.

\bibitem{tang2017depth}
Huixuan Tang, Scott Cohen, Brian Price, Stephen Schiller, and Kiriakos~N
  Kutulakos.
\newblock Depth from defocus in the wild.
\newblock In {\em Proceedings of the IEEE Conference on Computer Vision and
  Pattern Recognition}, pages 2740--2748, 2017.

\bibitem{trott2018interpretable}
Alexander Trott, Caiming Xiong, and Richard Socher.
\newblock Interpretable counting for visual question answering.
\newblock In {\em International Conference on Learning Representations}, 2018.

\bibitem{vu2018grounded}
Hoa~Trong Vu, Claudio Greco, Aliia Erofeeva, Somayeh Jafaritazehjani, Guido
  Linders, Marc Tanti, Alberto Testoni, Raffaella Bernardi, and Albert Gatt.
\newblock Grounded textual entailment.
\newblock In {\em Proceedings of the 27th International Conference on
  Computational Linguistics}, pages 2354--2368, 2018.

\bibitem{wang2017fvqa}
Peng Wang, Qi Wu, Chunhua Shen, Anthony Dick, and Anton Van Den~Hengel.
\newblock Fvqa: Fact-based visual question answering.
\newblock {\em IEEE transactions on pattern analysis and machine intelligence},
  40(10):2413--2427, 2017.

\bibitem{CAVE_0201}
M. Watanabe and S.K. Nayar.
\newblock {R}ational {F}ilters for {P}assive {D}epth from {D}efocus.
\newblock {\em International Journal on Computer Vision}, 27(3):203--225, May
  1998.

\bibitem{watanabe1996telecentric}
Masahiro Watanabe and Shree~K Nayar.
\newblock Telecentric optics for computational vision.
\newblock In {\em European Conference on Computer Vision}, pages 439--451.
  Springer, 1996.

\bibitem{willson1994modeling}
Reg~G Willson.
\newblock Modeling and calibration of automated zoom lenses.
\newblock In {\em Videometrics III}, volume 2350, pages 170--186. International
  Society for Optics and Photonics, 1994.

\bibitem{yang2020trrnet}
Xiaofeng Yang, Guosheng Lin, Fengmao Lv, and Fayao Liu.
\newblock Trrnet: Tiered relation reasoning for compositional visual question
  answering.
\newblock In {\em Computer Vision--ECCV 2020: 16th European Conference,
  Glasgow, UK, August 23--28, 2020, Proceedings, Part XXI 16}, pages 414--430.
  Springer, 2020.

\bibitem{yi2018neural}
Kexin Yi, Jiajun Wu, Chuang Gan, Antonio Torralba, Pushmeet Kohli, and
  Joshua~B. Tenenbaum.
\newblock Neural-symbolic vqa: Disentangling reasoning from vision and language
  understanding.
\newblock In {\em Advances in Neural Information Processing Systems}, pages
  1039--1050, 2018.

\bibitem{yu2016modeling}
Licheng Yu, Patrick Poirson, Shan Yang, Alexander~C Berg, and Tamara~L Berg.
\newblock Modeling context in referring expressions.
\newblock In {\em European Conference on Computer Vision}, pages 69--85.
  Springer, 2016.

\bibitem{yu2019deep}
Zhou Yu, Jun Yu, Yuhao Cui, Dacheng Tao, and Qi Tian.
\newblock Deep modular co-attention networks for visual question answering.
\newblock In {\em Proceedings of the IEEE/CVF Conference on Computer Vision and
  Pattern Recognition}, pages 6281--6290, 2019.

\bibitem{zellers2019recognition}
Rowan Zellers, Yonatan Bisk, Ali Farhadi, and Yejin Choi.
\newblock From recognition to cognition: Visual commonsense reasoning.
\newblock In {\em Proceedings of the IEEE/CVF Conference on Computer Vision and
  Pattern Recognition}, pages 6720--6731, 2019.

\bibitem{zhang2017ppr}
Hanwang Zhang, Zawlin Kyaw, Jinyang Yu, and Shih-Fu Chang.
\newblock Ppr-fcn: Weakly supervised visual relation detection via parallel
  pairwise r-fcn.
\newblock In {\em Proceedings of the IEEE International Conference on Computer
  Vision}, pages 4233--4241, 2017.

\bibitem{zhao2018semantically}
Handong Zhao, Quanfu Fan, Dan Gutfreund, and Yun Fu.
\newblock Semantically guided visual question answering.
\newblock In {\em 2018 IEEE Winter Conference on Applications of Computer
  Vision (WACV)}, pages 1852--1860. IEEE, 2018.

\bibitem{zhou2016learning}
Bolei Zhou, Aditya Khosla, Agata Lapedriza, Aude Oliva, and Antonio Torralba.
\newblock Learning deep features for discriminative localization.
\newblock In {\em Proceedings of the IEEE conference on computer vision and
  pattern recognition}, pages 2921--2929, 2016.

\bibitem{zhou2011coded}
Changyin Zhou, Stephen Lin, and Shree~K Nayar.
\newblock Coded aperture pairs for depth from defocus and defocus deblurring.
\newblock {\em International journal of computer vision}, 93(1):53--72, 2011.

\end{thebibliography}
